\documentclass[runningheads]{llncs}

\usepackage{eccv}

\usepackage{eccvabbrv}

\usepackage{graphicx}
\usepackage{booktabs, wrapfig}
\usepackage{subcaption}
\usepackage{enumitem}
\usepackage{multirow}
\usepackage{etoolbox}

\usepackage[accsupp]{axessibility}  

\usepackage[pagebackref,breaklinks,colorlinks,citecolor=eccvblue]{hyperref}

\usepackage{orcidlink}

\AtBeginDocument{
  \setlength{\parskip}{0pt plus 1pt minus 1pt}
}
\begin{document}

\title{Described Spatial-Temporal Video Detection} 

\titlerunning{Described Spatial-Temporal Video Detection}

\author{Wei Ji\inst{1} \and
Xiangyan Liu\inst{1} \and
Yingfei Sun\inst{1} \and
Jiajun Deng\inst{2} \and
You Qin\inst{1} \and
Ammar Nuwanna\inst{1} \and
Mengyao Qiu\inst{3} \and
Lina Wei\inst{3} \and
Roger Zimmermann\inst{1}
}

\authorrunning{Wei, Ji et al.}

\institute{National University of Singapore, Singapore \and University of Adelaide, Australia
\and Hangzhou City University, China
\\
\email{weiji0523@gmail.com, liu.xiangyan@u.nus.edu, sunyingfei@u.nus.edu}\\}

\maketitle

\begin{abstract}
    Detecting visual content on language expression has become an emerging topic in the community.
    However, in the video domain, the existing setting, \emph{i.e.}, \texttt{spatial-temporal video grounding} (STVG), is formulated to only detect one pre-existing object in each frame, ignoring the fact that language descriptions can involve none or multiple entities within a video.
    In this work, we advance the STVG to a more practical setting called \texttt{described spatial-temporal video detection} (DSTVD) by overcoming the above limitation.
    To facilitate the exploration of DSTVD, we first introduce a new benchmark, namely DVD-ST.
    Notably, DVD-ST supports grounding from none to many objects onto the video in response to queries and encompasses a diverse range of over 150 entities, including appearance, actions, locations, and interactions. 
    The extensive breadth and diversity of the DVD-ST dataset make it an exemplary testbed for the investigation of DSTVD.
    In addition to the new benchmark, we further present two baseline methods for our proposed DSTVD task by extending two representative STVG models, \emph{i.e.}, TubeDETR, and STCAT.
    These extended models capitalize on tubelet queries to localize and track referred objects across the video sequence.
    Besides, we adjust the training objectives of these models to optimize spatial and temporal localization accuracy and multi-class classification capabilities. 
    Furthermore, we benchmark the baselines on the introduced DVD-ST dataset and conduct extensive experimental analysis to guide future investigation. Our code and benchmark will be publicly available.
\end{abstract}

\section{Introduction}
\begin{figure}[!t]
    \centering
    \begin{subfigure}[b]{0.5\textwidth}
        \includegraphics[width=\textwidth]{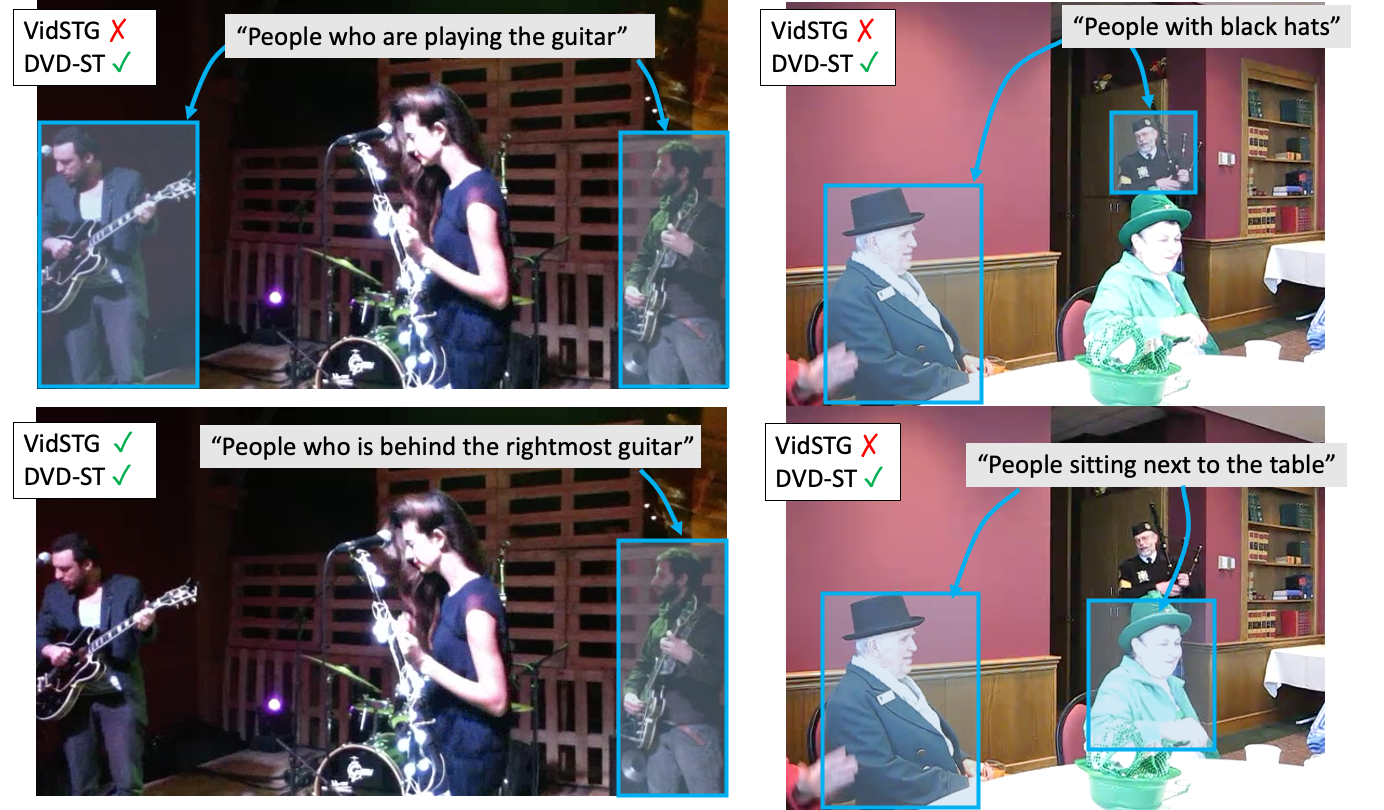}
        \caption{The VidSTG~\cite{zhang2020does} only refers to one object in a description, which leads to the lack of generalizability and limits the type of queries.}
        \label{fig:intro_a}
    \end{subfigure}%
    ~ 
    \begin{subfigure}[b]{0.5\textwidth}
        \includegraphics[width=\textwidth]{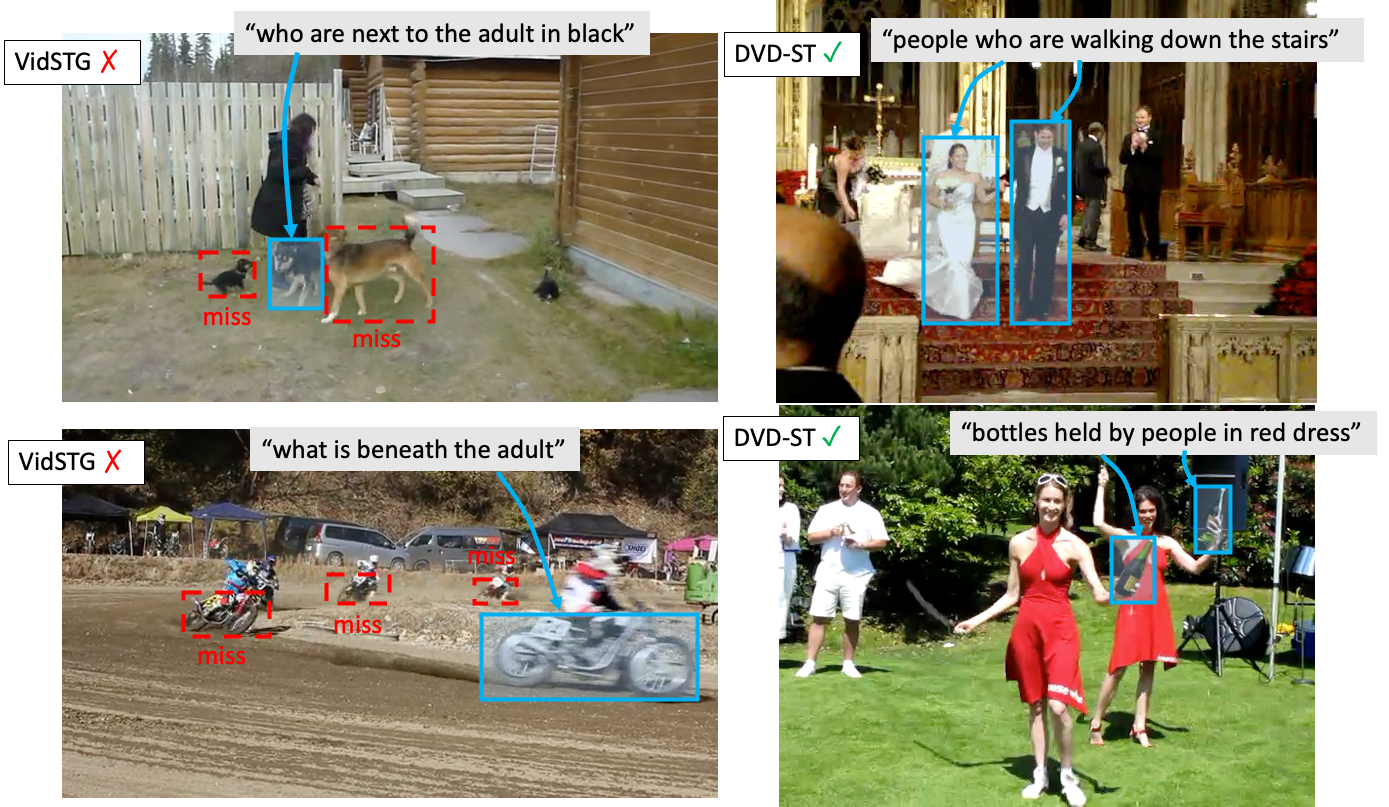}
        \caption{The VidSTG~\cite{zhang2020does} misses the annotation of some referred objects when one text query refers to multiple instances. Our DVD-ST is well-annotated to include all of the referred objects.}
        \label{fig:intro_b}
    \end{subfigure}
    \caption{\textbf{Comparison between the VidSTG dataset and our DVD-ST in terms of generalizability of descriptions and number of referred objects.} VidSTG is one of the representative STVG datasets, while our DVD-ST aims to benchmark a more practical described spatial-temporal video detection setting.}
    \label{fig:intro}
    \vspace{-2.em}
\end{figure}

The field of spatial-temporal video understanding \cite{yang2022tubedetr, jin2022embracing, wang2023efficient} has become increasingly significant in modern computer vision, with applications ranging from surveillance to interactive media \cite{vishwakarma2013survey, lin2023univtg}.
The rapid development of video technology and its immense application value have garnered significant attention for recommendation system \cite{ni2023content}, video content retrieval and localization \cite{Wang_2023_ICCV,Yan_2023_ICCV}. Language descriptions, as the most natural mode of human-computer interaction, are anticipated to serve as the query for video content detection.



In the literature, the task of detecting video content based on language expression is formulated as spatial-temporal video grounding (STVG) \cite{tang2021human,zhang2020does,yamaguchi2017spatio}.
As depicted in Figure~\ref{fig:intro}, STVG traditionally focuses on identifying a single instance as referred to by a text query.
However, real-world applications frequently require the analysis of multiple instances simultaneously, which presents a more complex challenge.
For instance, in surveillance and security, the ability to track multiple individuals or objects displaying suspicious behavior is crucial.
Similarly, in sports analysis, understanding the dynamics of a team often involves observing the coordinated movements of several players.
In traffic management, the effective monitoring of congested areas depends on the simultaneous tracking of multiple vehicles.
These scenarios, along with others such as crowd management at large events and consumer behavior analysis in retail environments, underscore the need for STVG algorithms that can adeptly handle multiple instances.
Moreover, the reality that sometimes no relevant content is found in the video sequence further complicates the task.
These practical challenges highlight the limitations of current STVG benchmarks and the urgent need for the development of more sophisticated algorithms that can cater to the diverse and complex demands of language-based video content detection in real-life applications.

To this end, we propose a more practical setting, namely, described spatial-temporal video detection (DSTVD),  complemented by the introduction of a new benchmark, DVD-ST.
As illustrated in Figure~\ref{fig:intro}, DVD-ST is characterized by its distinctive attributes:
\textbf{1) Multiple object grounding:} Diverging from conventional STVG datasets, DVD-ST associates one text query with a varying number of objects;
\textbf{2) Diverse entities:} The benchmark encompasses queries referring to a rich tapestry of over 150 entities, spanning appearances, actions, locations, and interactions. This provides a breadth and depth of content that surpasses existing datasets;
\textbf{3) Instance-level annotations:} DVD-ST offers comprehensive instance-level annotations, facilitating detailed video content analyses. \looseness=-1

Besides, we have adapted two existing STVG models, TubeDETR \cite{yang2022tubedetr} and STCAT \cite{jin2022embracing}, to align with the nuanced requirements of DSTVD.
These adaptations enhance the models' ability to interpret complex text queries and accurately localize multiple objects in video sequences.
Key modifications include the integration of tubelet queries and a tubelet-wise matcher, essential for tracking multiple objects in accordance with text queries.
Furthermore, we refined the training objectives, focusing on spatial-temporal localization accuracy and multi-class classification, to equip the models for the specific challenges of DSTVD. \looseness=-1

\textbf{Our key contributions are summarized as follows:} \vspace{-0.5em}
\begin{enumerate}
    \item We pioneer the concept of described spatial-temporal video detection (DSTVD), marking a paradigm shift toward a more generalized and inclusive form of video content detection on language descriptions.

    \item We introduce DVD-ST, a rigorously curated video understanding benchmark with manual annotations to foster research on temporal action reasoning. Through comprehensive analysis, we benchmark existing spatio-temporal video reasoning techniques on our dataset. 
    
    \item We have modified and augmented the TubeDETR and STCAT frameworks to be tailored for the DSTVD task, enabling these models to effectively handle more complex, real-world scenarios involving multiple objects and diverse queries. We also provide robust evaluation metrics, ensuring a thorough and detailed approach to assessing performance in DSTVD tasks.
\end{enumerate}

\section{Related Works} \label{sec:related_works}

\subsection{Datasets}
\paragraph{Spatial-Temporal Detection Dataset.}
Video action detection and video object detection represent crucial tasks in the domain of spatial-temporal detection.
The former, exemplified by datasets like VIRAT \cite{oh2011large}, AVA \cite{gu2018ava}, UCF Sports \cite{rodriguez2008action}, and MAMA \cite{modi2022video}, focuses on identifying human-centric activities in video content.
In contrast, video object detection, primarily evaluated on the widely used ImageNet VID dataset \cite{russakovsky2015imagenet}, is concerned with recognizing and localizing objects in the spatial-temporal context of video frames. 
It is important to note that datasets for video action detection and video object detection lack corresponding textual annotations.

\paragraph{Spatial-Temporal Video Grounding Dataset.} 
Spatial-temporal video grounding (STVG) is a pivotal task that involves identifying and localizing a referred region within a video based on a given textual description. This task is particularly challenging due to the dynamic nature of videos and the complexity of natural language.
An array of datasets have been collected to benchmark STVG, including ActivityNet-SRL \cite{sadhu2020video}, VidSTG \cite{zhang2020does}, HC-STVG \cite{tang2021human}, VID-sentence \cite{chen2019weakly}, and GroundingYouTube \cite{chen2023and}.
Notably, HC-STVG and STPR \cite{yamaguchi2017spatio} primarily focus on a human-centric perspective, with ActivityNet-SRL and STPR being derived from ActivityNet \cite{caba2015activitynet}.
VidSTG, on the other hand, originates from VidOR \cite{shang2019annotating}, where the authors extended the original dataset by annotating different sentence structures such as declarative and interrogative sentences.
However, these datasets share a common characteristic—they adhere to a paradigm where a textual description corresponds to an object in the video. This somewhat limits the practical applicability of spatial-temporal video grounding in real-world production environments.

\paragraph{Other Related Datasets.} 
While several related tasks and benchmarks exist, they differ significantly from DSTVD.
In Referring Multi-object Tracking (RMOT), although each text description may correspond to multiple objects in a video, the primary focus remains on tracking. Notably, RMOT's representative dataset, Refer-KITTI \cite{wu2023referring}, is comparatively smaller in dataset size and features less intricate text descriptions than DVD-ST, owing to inherent task disparities.
Additionally, in image-level tasks like Referring Image Segmentation and Referring Image Comprehension, there are parallels, yet our video-level task inherently introduces heightened complexity in the temporal dimension, presenting more formidable challenges (refer to Section \ref{subsec:definition}).

\subsection{Methods}
Given the characteristics of the DSTVD task, two areas of work are highly relevant at the technical solution level.
The first direction involves transformer-based object detection \cite{detr, kamath2021mdetr, yang2022tubedetr, zhu2020deformable}, where objects are decoded in parallel at each transformer decoder layer, eliminating the need for any prior design and achieving end-to-end modeling.
The second direction is spatial-temporal video grounding; considering the similarity between its input and output with DSTVD, frameworks for such tasks can serve as the backbone for addressing DSTVD. As pioneers introducing the new DSTVD task, we will judiciously integrate these two directions to provide a simple and effective solution for DSTVD.

\section{Benchmark} \label{sec:new_benchmark}

\subsection{Task Setting} \label{subsec:definition}
\paragraph{Problem Formulation.} Given the unrestricted text description, DSTVD aims to identify all objects referenced in the video, providing both temporal and spatial localization for the targeted subjects.
This type of video detection considers not only the spatial characteristics of objects (their appearance, shape, and location in individual frames) but also their temporal aspects (how these objects or events change and move over time).
The input to this task is a video $V$ and a textual description $D_t$, and the output is a series of bounding boxes $B$, each series corresponding to a referred object.

To mathematically formulate the DSTVD task, we consider the following: The textual description, represented as $D_t$, provides the narrative or keywords to identify the referred objects within the video.
The video input is denoted as $V = \{ F_1, F_2, ..., F_n \}$, with each $F_i$ representing a frame in the video.
Each object $O_j$ referred to in the textual description $D_t$ is identified.
For each referred object $O_j$ in a frame $F_i$, a bounding box $B_{ij}$ is defined, representing the spatial localization of the object in that frame.
The detection function $Det(V, D_t)$ processes the video $V$ and the textual description $D_t$ to output a set of bounding boxes for each referred object.
Formally, $Det: (V, D_t) \rightarrow \{ B_{ij} \}$, where $B_{ij}$ is the bounding box for referred object $O_j$ in frame $F_i$.
The output is a series of bounding boxes $\{ B_{1j}, B_{2j}, ..., B_{nj} \}$ for each referred object $O_j$ identified from the textual description $D_t$.
These bounding boxes provide the spatial and temporal localization of the object throughout the video.
It is important to highlight that the count of referred objects in DSTVD can range from zero to any number, accommodating a wide spectrum of scenarios.

In summary, the DSTVD task is defined mathematically as finding the function $Det(V, D_t)$ that maps a video $V$ and a textual description $D_t$ to a series of bounding boxes $B$, each series tracking a referred object $O_j$ across the video frames. This approach integrates both spatial and temporal dimensions, leveraging textual descriptions to guide the identification and tracking of referred objects within dynamic video content.

\paragraph{Key Challenges.}
To facilitate a deeper understanding of the new DSTVD benchmark, and to delineate its unique contributions compared to the benchmarks in the literature, we enumerate the principal challenges as follows:
\begin{enumerate}
    \item \textbf{Arbitrary number of referred objects.} In our DSTVD benchmark, a text query's reference to objects is variable—ranging from none to one or multiple objects. Additionally, the number of referred objects for the same video and text query can vary across different frames.
    \item \textbf{Temporal consistency in the tubelet.} 
    In the same video, it is crucial to differentiate boxes predicted in different frames belonging to the same tubelet. The final prediction output does not consist of individual boxes for each frame; rather, it comprises tubelets representing all referenced objects in the video. Specifically, each tubelet has a predicted box in the corresponding frame, all pertaining to the same object.
\end{enumerate}

\begin{figure}[t]
    \centering
    \begin{minipage}{0.48\textwidth}
        \centering
        \includegraphics[width=\textwidth]{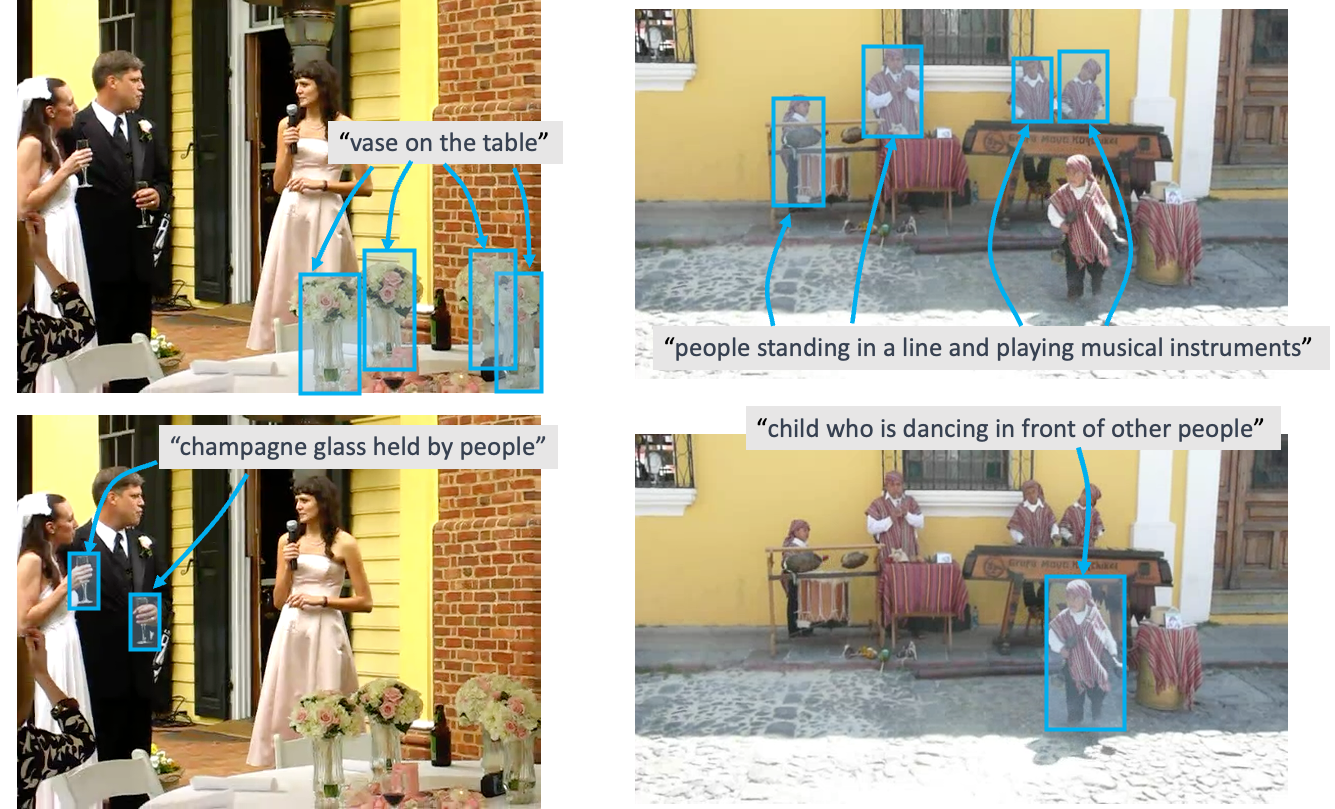}
        \caption{Examples of the described queries from DVD-ST, which have abundant entities in semantics.}
        \label{fig:examples}
    \end{minipage}\hfill
    \begin{minipage}{0.48\textwidth}
        \centering
        \includegraphics[width=\textwidth]{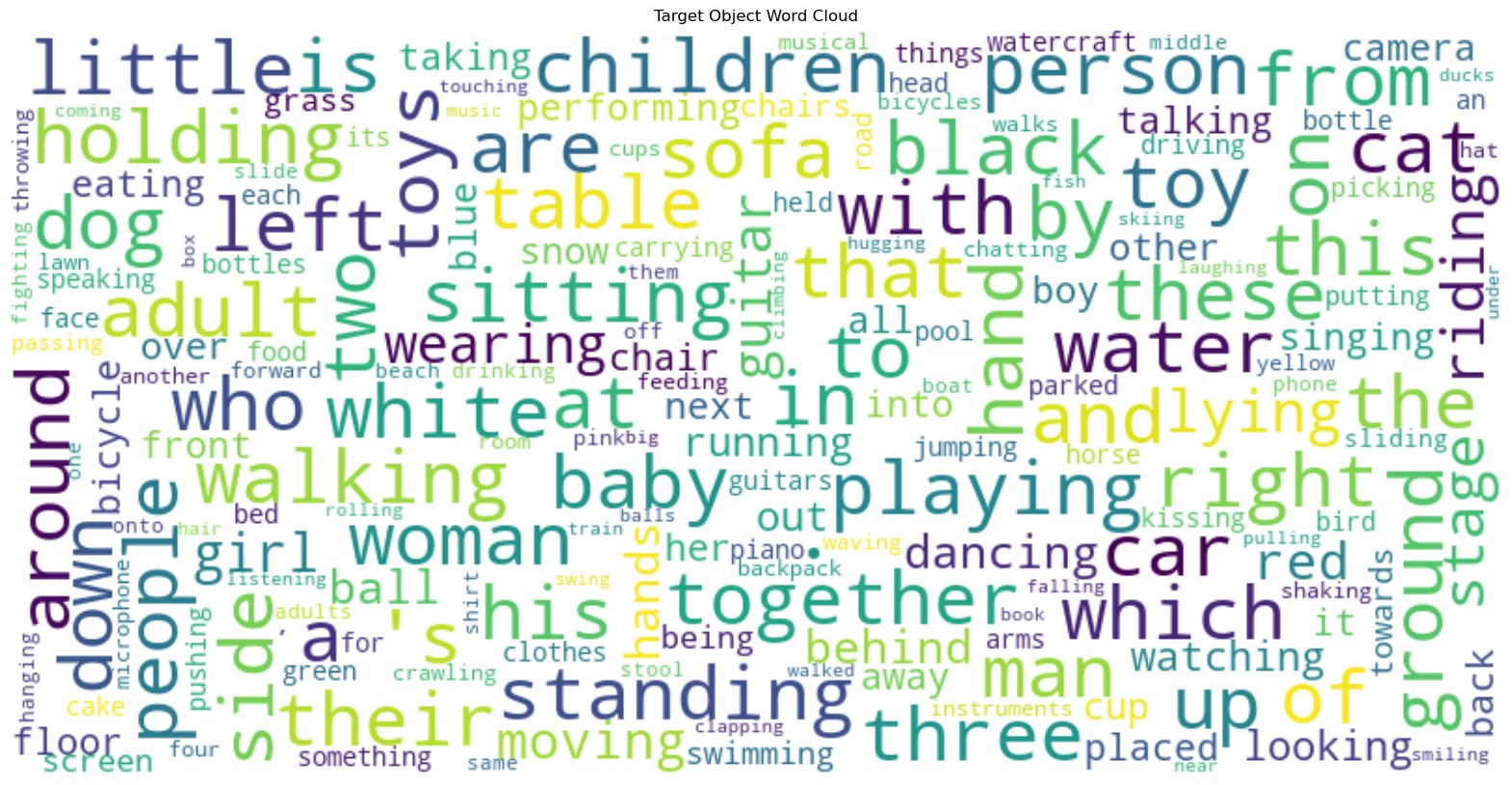}
        \caption{Word cloud of the described queries from DVD-ST, which includes sufficient object and relation entities.}
        \label{fig:word_cloud}
    \end{minipage}
    \vspace{-1.em}
\end{figure}

\subsection{Data Collection and Annotation}\label{subsec:data}

\paragraph{Source dataset collection.}
Our primary video source is the VidOR dataset \cite{shang2019annotating}, which includes original annotations, bounding boxes for each object, and object types within the videos.
The VidOR dataset \cite{shang2019annotating} was selected due to its diverse range of scenes and the frequent presence of multiple objects of the same type, showcasing a variety of actions, appearances, and characteristics.

\paragraph{Image caption collection.}
For frame-specific captioning, we utilize the Vit-GPT-image-caption model \cite{nlp_connect_2022}.
It is important to note that these captions are not directly used as queries for our DVD-ST; instead, they serve to aid annotators in comprehending the overall content of the frames.

\paragraph{Description generation and instance-level annotation.}
Employing image captions as navigational aids, our approach involves meticulously creating descriptions that encompass multiple target objects, coupled with annotating their corresponding start and end frames within the videos.
To streamline the annotation workflow, we capitalize on existing data, such as object indices and categories in each frame, and employ a custom-designed, efficient labeling tool.
This process, exemplified in Figure \ref{fig:dataset_statistics} (a), includes the display of all bounding boxes during the annotation phase, enhancing precision and efficiency.

\subsection{Dataset Statistics}\label{subsec:statistics}
\begin{table*}[t]
    \centering
    \caption{Statistics of Spatio-Temporal Video Grounding Datasets.}
    \vspace{-1.em}
    \scalebox{0.9}{\begin{tabular}{lcccccc}
        \toprule
        & STPR ~\cite{yamaguchi2017spatio} & VID-Sentence \cite{chen2019weakly}  & VidSTG \cite{zhang2020does}  & HC-STVG \cite{tang2021human} & DVD-ST \\
        \midrule
        Source & ActivityNet & VID & YFCC100M & AVA & YFCC100M \\
        \# Descriptions & 30,365 & 7,654 & 99,943 & 5,660 & 5,734 \\
        \# Videos & 5,293 & 5,318 & 6,924 & 5,660 & 2,750 \\
        \# Object Categories & 1 & 30 & 79 & 1 & 163 \\
        Single-Object & $\checkmark$ & $\checkmark$ & $\checkmark$ & $\checkmark$ & $\checkmark$ \\
        Multi-Object & $\times$ & $\times$ & $\times$ & $\times$ & $\checkmark$ \\
        \bottomrule
    \end{tabular}}
    \label{tab:dataset-comparison}
    \vspace{-2.em}
\end{table*}

For our dataset, we have compiled a total of 5734 descriptions across 2750 annotated videos, averaging 2.08 descriptions per video.
A comparative overview with other existing spatio-temporal video grounding datasets is presented in Table~\ref{tab:dataset-comparison}.
We have segmented the dataset into three distinct sections: training, validation, and testing, comprising 1699, 421, and 632 videos, and corresponding to 3114, 1293, and 1327 descriptions, respectively.
The range of target objects per query varies from 1 to 12, with an average of 1.81 target objects.
Additionally, the average query length in our dataset is 7.54 words.
Although the DVD-ST dataset does not surpass previous STVG datasets in terms of absolute numbers (videos and descriptions), it stands out in its annotation complexity and the diversity of object types involved.
Moreover, it serves as a pioneering dataset for the DSTVD task.

\begin{figure*}[!t]
    \centering
    \begin{subfigure}[b]{0.32\textwidth}
        \includegraphics[width=\linewidth]{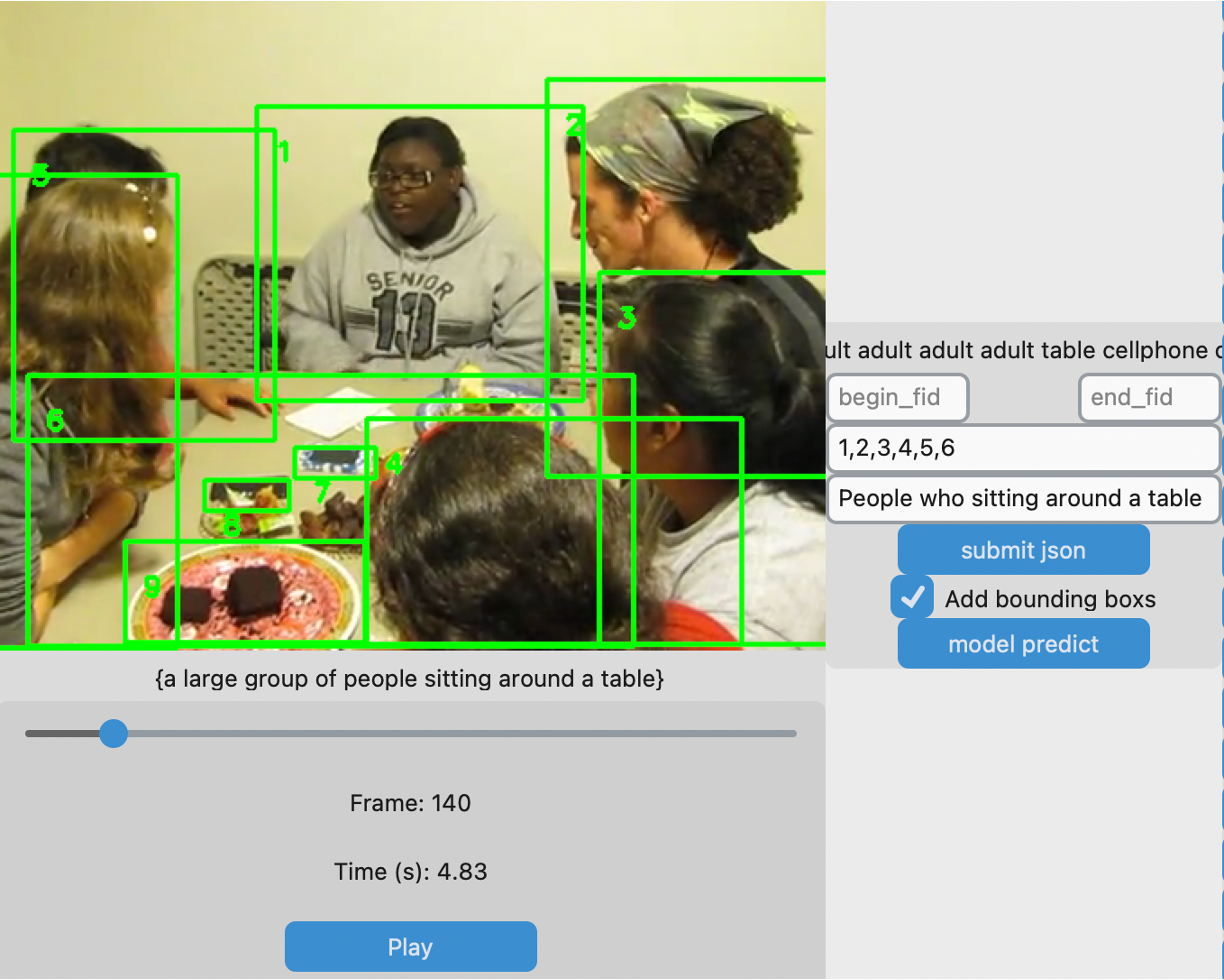}
        \caption{Surface of the self-developed annotation platform.}
        \label{fig:annotation_process}
    \end{subfigure}\hfill 
    \begin{subfigure}[b]{0.32\textwidth}
        \includegraphics[width=\linewidth]{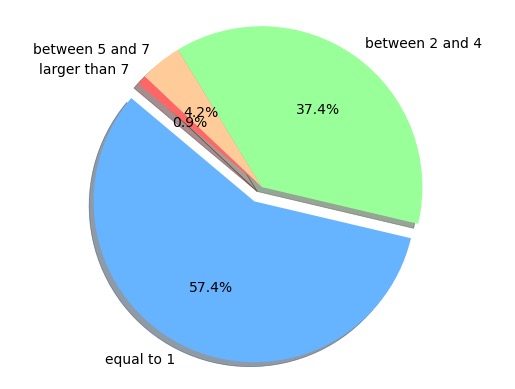}
        \caption{Distribution of the target objects count.}
        \label{fig:object_list}
    \end{subfigure}\hfill 
    \begin{subfigure}[b]{0.32\textwidth}
        \includegraphics[width=\linewidth]{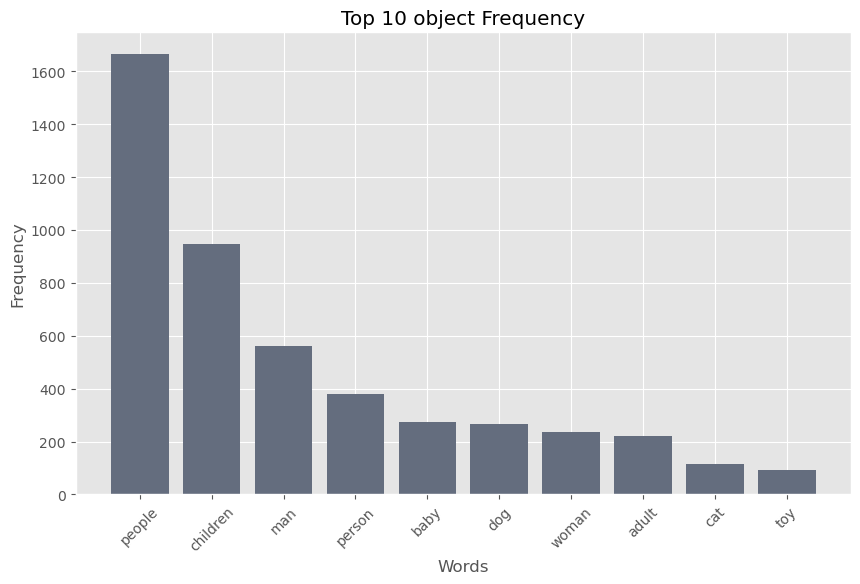}
        \caption{Top 10 frequent target objects.}
        \label{fig:top_10}
    \end{subfigure}
    \caption{\textbf{Overview of the annotation platform and dataset statistics:} (a) shows the interface of the annotation platform, (b) illustrates the distribution of objects, and (c) presents the most frequent objects in the dataset.}
    \vspace{-2.em}
    \label{fig:dataset_statistics}
\end{figure*}

We further analyzed the DVD-ST dataset, revealing in Figure \ref{fig:dataset_statistics} (b) that about half of the descriptions involve non-single-object annotations.
Figure \ref{fig:dataset_statistics} (c) shows that the most frequently annotated subjects are predominantly human characters, influenced by the original VidOR dataset and the action-focused nature of the descriptions.
This analysis also highlights the presence of other commonly annotated subjects like animals and inanimate objects, adding to the dataset's diversity

\subsection{Evaluation Metrics} \label{subsec:metric}
In our benchmark, we propose to evaluate the performance of algorithms for described spatial-temporal video detection by considering the capability of spatial, temporal, and spatial-temporal localization. Specifically, we exploit \texttt{m\_vIoU}, \texttt{tIoU}, and \texttt{vIoU@R} to evaluate the accuracy of the temporal localization and spatial localization of objects.
Moreover, since one text description can refer to multiple instances, we further introduce \texttt{frame-AP} and \texttt{video-AP} to judge whether a model can distinguish different referred instances. The details of these metrics are listed as follows:


\textbf{Spatial} (\texttt{m\_vIoU}, \texttt{vIoU@R}). From a spatial viewpoint, it's essential to measure the accuracy of the model's predictions regarding the bounding boxes of detected objects. Following \cite{yang2022tubedetr, jin2022embracing}, we define \texttt{vIoU} to assess the model's match with each ground truth tubelet: \texttt{vIoU}$=\frac{1}{S_u}\sum_{t \in S_i}IoU(\hat{b}_t, b_t)$, where $S_i$ and $S_u$ are the set of frames in the intersection and union respectively, $\hat{b}_t$ and $b_t$ respectively denote the predicted box and the ground truth box at time $t$.
Moreover, we introduce \texttt{m\_vIoU}, the average IoU across all tubelets, to evaluate the model's overall predictions for tubelets: \texttt{m\_vIoU}$=\frac{1}{\sum_{j=1}^N n_j}\sum_{j=1}^N\sum_{k=1}^{n_j} vIoU_j^k$, where $N$ denotes the number of samples, and $n_j$ means the number of tubelets in the $j$-th sample.
Additionally, for a more fine-grained evaluation, we also introduce \texttt{vIoU@R}, which represents the proportion of samples for which \texttt{vIoU} is greater than \texttt{R}.

\textbf{Temporal} (\texttt{tIoU}). From a temporal analysis standpoint, accurately evaluating the initial frames of detected tubelets in the model's predictions is essential. To facilitate this assessment, we introduce the temporal Intersection over Union (\texttt{tIoU}) metric. This metric is defined as \texttt{tIoU} = $\frac{1}{\sum_{j=1}^N n_j}\sum_{j=1}^N\sum_{k=1}^{n_j}\frac{S_i(j,k)}{S_u(j,k)}$, and is specifically designed to gauge the model's performance in temporal localization. Here, $S_i(j,k)$ denotes the intersection score ($S_i$) for the $k$-th tubelet in the $j$-th sample, and $S_u(j,k)$ denotes the union score ($S_u$) for the $k$-th tubelet in the $j$-th sample

\textbf{Multi-class Classification} (\texttt{frame-AP}, \texttt{video-AP}). Additionally, we also incorporated two metrics, \texttt{frame-AP} and \texttt{video-AP}, inspired by \cite{gkioxari2015finding}, to assess the model's performance at both frame and video levels in terms of classification. 
\texttt{frame-AP} and \texttt{video-AP} calculate the area under the precision-recall curve for detection in each frame and the action tube predictions, respectively. 
In \texttt{frame-AP@R}, a detection is considered correct if the $IoU$ with the ground truth at that frame exceeds the threshold \texttt{R}. 
For \texttt{video-AP@R}, a tube is deemed correct if the average per-frame $IoU$ with the ground truth across all frames of the video surpasses the threshold \texttt{R}.

\subsection{Annotation Quality Control}\label{subsec:annotation_control}

The annotation process for the DVD-ST dataset was meticulously conducted over a period of four months by a team of six undergraduate students, organized into three distinct stages. To guarantee the quality of annotations, we adhered to several guiding principles:

\begin{itemize}[label=\textbullet, leftmargin=12pt, nosep]
    \item \textbf{Pre-annotation training:} Each annotator underwent rigorous training before commencing actual annotation work, ensuring a standardized understanding of the task requirements.
    \item \textbf{Tool-assisted annotation:} The annotation process was facilitated by a captioning model, as detailed in Section \ref{subsec:data}. This approach, coupled with our specially designed annotation tool, significantly enhanced the accuracy of time-stamped annotations and the quality of descriptive generation.
    \item \textbf{Length control and focus shift:} We set a recommended maximum length for the target object list at 15 items, particularly advocating concise annotations in videos featuring numerous entities. To maintain a balanced distribution of target object quantities, we continuously monitored and analyzed the annotated data. This allowed us to shift our focus appropriately---from single object descriptions to multi-object descriptions---once a sufficient quantity of the former was achieved.
    \item \textbf{Quality control for video selection:} Annotators were encouraged to report videos that posed challenges for effective description, such as those with a scarcity of describable entities. Videos confirmed as lacking in descriptive potential were subsequently removed from the dataset, ensuring the overall quality and relevance of the database.
\end{itemize}

These structured approaches were instrumental in maintaining high standards throughout the annotation process, ensuring that the DVD-ST dataset was annotated with both precision and relevance.

\subsection{Dataset Highlight}\label{subsec:dataset_highlight}
Our re-annotation of the DVD-ST dataset, based on VidOR \cite{shang2019annotating}, aims to enhance understanding of video object relations.
As illustrated in Figure \ref{fig:intro}, we compare examples from VidSTG, also re-annotated on VidOR, with our dataset to highlight DVD-ST's unique characteristics.
The key features of our dataset include a focus on videos with multiple objects of the same type, such as groups of people, toys, or cars. This approach differs from earlier spatial-temporal video grounding datasets, which concentrate on single object descriptions.
Our challenge involves designing queries that are sufficiently general to accurately refer to any number of objects, while remaining contextually relevant to the video content.

Additionally, our dataset is characterized by a rich variety of query types. Annotations consider multiple aspects, including appearance, actions, location, and interactions. This diversity, showcased in Figure \ref{fig:examples}, contributes to the depth and variety of our annotated statements. The video content spans various scenes, allowing our queries to encompass a wide range of objects and scenarios, as depicted in the word cloud in Figure \ref{fig:word_cloud}. \looseness=-1

Another distinguishing feature is our commitment to instance-level annotation. We manually annotate the start and end frames for each query to cater to instances where the object of interest appears multiple times. Any frame containing the queried object is thus deemed relevant. This meticulous approach underlines the comprehensive and detailed nature of the DVD-ST dataset, designed to provide an extensive understanding of video object relations.

\section{Method}
\label{sec:method}
To investigate the performance of existing methods on the DSTVD task, we selected two representative frameworks: TubeDETR~\cite{yang2022tubedetr} and STCAT~\cite{jin2022embracing}, both within the domain of Spatial-Temporal Video Grounding (STVG), a task closely related to our objective at the methodological level. We made adaptive improvements based on the selected frameworks to be able to solve the DSTVD task.

\begin{figure*}[!t]
    \centering
    \includegraphics[width=\textwidth]{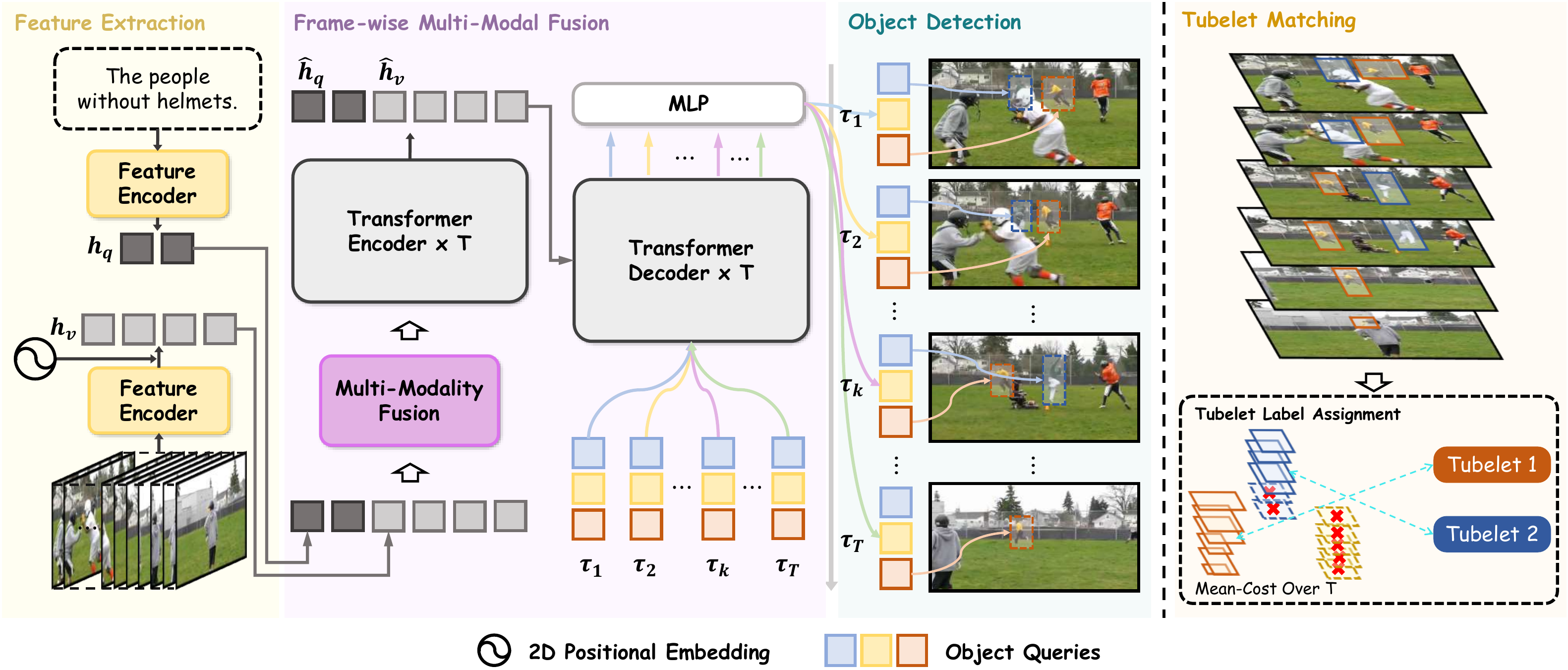}
    \caption{\textbf{Illustration of our proposed TubeDETR-M framework, which is a simple yet effective baseline for DSTVD task.} All input video frames and the description are first processed with a Visual Encoder and a Text Encoder. The resulting text $ h_v$ and video features $ h_q$ are then jointly encoded with a Video-Text Encoder that computes spatial and multi-modal interactions. The resulting video-text features are then decoded into the output spatio-temporal tube using a Transformer Decoder, which is guided by tubelet queries. Our adaptations for DSTVD primarily focus on 1) improvements to the decoder input side, and 2) the introduction of a tubelet-wise matcher. These enhancements align with our another framework, STCAT-M.
 }
    \vspace{-0.5cm}
    \label{fig:framework}
\end{figure*}

\subsection{STVG Frameworks}
Current STVG methods rely on a strong assumption that \emph{each text query for the video maps to a particular object's tubelet}.
Consequently, these methods can solve the STVG task relatively easily by identifying whether the target object exists in each video frame and predicting its bounding box.
By linking these predicted bounding boxes, the tubelet indicated by the query can be reconstructed naturally.

Among these methods, TubeDETR \cite{yang2022tubedetr} and STCAT \cite{jin2022embracing} stand out as representative frameworks. 
TubeDETR introduces a novel space-time transformer decoder, and a dual-stream encoder for efficient spatial and multi-modal interactions. STCAT addresses feature alignment and prediction inconsistencies ignored by existing methods. The following is a detailed introduction to the methodologies of the two methods:\looseness=-1

\textbf{Framework 1: TubeDETR \cite{yang2022tubedetr}.} 
TubeDETR's network architecture consists of an encoder and a decoder.
The encoder employs a dual-stream approach, efficiently capturing spatial and multi-modal interactions using a slow multi-modal stream and a lightweight fast visual stream.
In the decoder, inspired by DETR\cite{detr}'s modeling paradigm, time queries are utilized to interact with the multi-modal features generated by the encoder. 
This interaction enables the detection of the referenced target for each frame.

\textbf{Framework 2: STCAT \cite{jin2022embracing}.} 
STCAT's network architecture consists of an encoder and a decoder as well. The multimodal encoder consists of spatial interaction layer and temporal interaction layer to perform a more consistent feature alignment between visual feature and text feature. In the decoder, inspired by \cite{44,yang2020improving}, content queries and position queries are generated by a template generation
 module to correlate and restrict the predictions across all video frames. Then use query-guided decoder and a prediction head to generate the final prediction.

To adapt to our new benchmark and enable the two frameworks to handle references to any number of objects (rather than just a single object in STVG tasks), we 1) introduced tubelet queries on the decoder side in Section \ref{subsec:tubelet_queries} and 2) utilized a tubelet-wise matcher to achieve the detection of multiple tubelets in Section \ref{subsec:tubelet_wise_target_assignment}. Additionally, We 3) redesigned the training loss to adapt to our new task in Section \ref{subsec:training_objective}. In brief, we made foundational adjustments building upon existing works, serving as a starting point for the DSTVD task.

\subsection{Tubelet Queries} \label{subsec:tubelet_queries}
Due to the requirements of DSTVD, which entail identifying tubelets for multiple objects specified by a textual query, the conventional methodology employed in traditional STVG tasks, limited to locating a single tubelet, becomes ineffective in this context.
Drawing inspiration from the transformer-based object detection \cite{detr, kamath2021mdetr}, we introduced the concept of \emph{tubelet queries} to facilitate the localization of an arbitrary number of tubelets.

In detail, when analyzing each frame of the video, the model employs tubelet queries and the relevant temporal queries (e.g., time query for TubeDETR and position query for STCAT) to locate all the referred objects. 
The total number of identified objects is limited by the quantity of tubelet queries used.

Notably, in TubeDETR and STCAT, tubelet queries will continue to adhere to the mechanism of time-aligned cross-attention.
This ensures that each tubelet query in every frame only attends to the video-text features corresponding to that specific frame. 

The difference between the generation of tubelet queries in TubeDETR and STCAT is that Tubedetr generate tubelet queries randomly, while STCAT generate tubelet queries from the output of encoder, which is the combination of local frame embedding and global embedding. In STCAT, each tubelet query only learns a partial set of cross-modal feature information.
\subsection{Tubelet-wise Target Assignment} \label{subsec:tubelet_wise_target_assignment}
In the previous section, the inclusion of tubelet queries theoretically enables the model to detect an arbitrary number of target objects in every frame of the video.
However, if we adopt the conventional approach \cite{detr, kamath2021mdetr}, treating the relationships between frames as independent entities and directly matching predicted bounding boxes with their respective ground truth boxes for each frame, it results in the loss of tubelet-specific information. 
This means it becomes difficult to discern which predicted bounding box corresponds to a particular tubelet in each frame.

Based on this, for the new DSTVD task, we introduced an additional \emph{tubelet-wise matcher}. 
This matcher not only matches the ground truth boxes but also specifies which predicted bounding box corresponds to which tubelet.

To explain further, firstly, we computed the matching loss between the ground truth boxes of all existing objects in each frame and the predicted boxes. 
Subsequently, considering the matching loss across all frames, we computed the average loss for the box series of each ground truth tubelet and the corresponding box series of each tubelet query across all frames.
Then, employing the Hungarian algorithm, we completed the tubelet-wise matching, ensuring that each predicted box associated with a tubelet query across different frames belongs to the same tubelet. More details about the matching loss refer to the Supplementary Material. \looseness=-1

\subsection{Training Objective} \label{subsec:training_objective}
At the optimization stage, we aimed to reuse the optimization functions of the TubeDETR and STCAT frameworks as much as possible and make necessary adaptive adjustments tailored to the DSTVD task.

Specifically, for TubeDETR and STCAT, we retained the existing box loss, which involves calculating the generalized Intersection over Union (gIoU \cite{rezatofighi2019generalized}) and L1 loss between predicted boxes and ground truth boxes. 
We also kept the current guided attention loss.  Additionally, we made adjustments to the temporal loss in TubeDETR and STCAT.  
We modified it to predict the start and end frames for each tubelet individually, rather than providing predictions for the entire video's start and end frames.
Furthermore, we introduced a three-way classification loss: it classifies each box predicted by the tubelet query into one of three states—1) exists and is referenced, 2) exists but is not referenced, and 3) does not exist. \looseness=-1

\section{Experiments}
\label{sec:experiments}
In this section, we first showcase the performance of two new models introduced in Section \ref{subsec:tubelet_queries} on DSTVD. Details of the experiment implementation are discussed in Section \ref{subsec:implementation_details}, main experimental results, comprehensive discussions, and analysis are reported in Section \ref{subsec:experimental_results_and_analysis}.

\subsection{Implementation Details} \label{subsec:implementation_details}
\paragraph{Network Architecture.} Our experiments aim to explore how existing methods perform on the DSTVD task. Introducing additional modules for more intricate modeling is beyond our scope. Thus, we present two new frameworks, TubeDETR-M and STCAT-M, built upon the TubeDETR and STCAT networks. The incorporation of new modules (details in Sec \ref{sec:method}) is to make existing models adaptable to our DSTVD task, with no changes to the backbone network structure. Both frameworks employ pre-trained ResNet-101 \cite{he2016deep} as the visual encoder and RoBERTa \cite{liu2019roberta} as the text encoder. \looseness=-1

\paragraph{Training and Inference.}
Given the constraints posed by hardware limitations and the importance of maintaining code cleanliness, we opted for a batch size of 1 throughout our experiments.
Furthermore, since our paper does not primarily aim for superior experimental results, our approach was to closely adhere to the original paper's hyperparameter settings without doing hyperparameter searching.
Additionally, for the newly introduced components, tubelet queries and loss, we set the number of tubelet queries to 15 for TubeDETR and 12 for STCAT assigned a weight to the loss function of 3 in both TubeDETR-M and STCAT-M.

\paragraph{Others.} 
During the inference stage, we utilized the tubelet-wise matcher to predict match predicted tubelets with ground truth tubelets. This approach aids in examining the model's performance in spatial and temporal dimensions through IoU-related metrics. For evaluating the model's classification performance, we can focus on AP-related metrics.

\subsection{Experimental Results and Analysis} \label{subsec:experimental_results_and_analysis}
Table \ref{tab:main_table} presents results for the TubeDETR-M and STCAT-M frameworks across different distributions of referenced object quantities. It's important to note that in this context, the models are trained on the complete training dataset. Table \ref{tab:single_object} compares STCAT-M with the original STCAT in \emph{single-object} scenarios (using the default STVG setting). The goal is to investigate whether the introduction of our additional design compromises the effectiveness of TubeDETR and STCAT in \emph{single-object} situations. Importantly, in this case, the models are trained only on the \emph{single-object} training dataset. Tables \ref{tab:description_length} and \ref{tab:entity_counts} demonstrate how STCAT-M performs in experiments with varying levels of query complexity. In Table \ref{tab:description_length}, complexity is assessed based on query length, with longer queries considered more complex. In Table \ref{tab:entity_counts}, complexity is determined by the number of entities mentioned in the query, where a higher entity count indicates greater complexity.

\begin{table*}[t]
    \centering
    \caption{Performance comparison of two baselines on our DVD-ST benchmark.}
    \vspace{-1.em}
    \label{tab:main_table}
    \scalebox{0.9}{
        \begin{tabular}{llcccccc}
        \toprule
            \textbf{Framework} & \textbf{Category}  & \texttt{m\_vIoU} & \texttt{tIoU} & \texttt{vIoU@0.3} & \texttt{vIoU@0.5} & \texttt{frame-AP@0.5} & \texttt{video-AP@0.25} \\ 
            \midrule
            \multirow{3}*{\textbf{TubeDETR-M}} & \emph{single-object} & 24.03 & 50.57 & 31.66 & 17.63 & 51.06 & 32.21\\ 
            ~ & \emph{multi-objects} &  65.15 & 70.70 & 63.82 & 51.32 & 29.23 & 53.85\\
            ~ & \emph{full} &  52.29 & 64.40 & 53.84 & 40.88 & 41.15 & 39.94 \\
            \midrule
            \multirow{3}*{\textbf{STCAT-M}} & \emph{single-object} & 40.56& 38.51 &
            58.08 & 46.25& 53.86& 61.26 \\
            ~ & \emph{multi-objects} &  33.26 &22.14 &27.35&24.81&40.75&14.81\\
            ~ & \emph{full} & 35.64 & 27.48& 37.38& 31.80  &47.18 & 42.32\\
            \bottomrule
        \end{tabular}
    }
    \label{exp:main_table}
    \vspace{-1.5em}
\end{table*}

\paragraph{Comparison across different object counts.}
Table \ref{tab:main_table} displays the performance of our improved TubeDETR-M and STCAT-M on the DSTVD task using the DVD-ST dataset, based on TubeDETR and STCAT. The results indicate that on the \emph{full} test set, TubeDETR-M outperforms STCAT-M in matching (IoU) metrics but performs poorly in classification (AP) metrics. Furthermore, on the \emph{single-object} test set, STCAT-M exhibits overall better performance compared to TubeDETR-M. However, it's worth noting that TubeDETR-M achieves a higher \texttt{tIoU} (50.57) on the \emph{single-object} test set than STCAT-M (38.51), indicating more accurate temporal predictions by TubeDETR-M. Then, on the \emph{multi-object} test set, TubeDETR-M outperforms STCAT-M across all evaluation metrics. Additionally, for \texttt{frame-AP@0.5}, where \emph{multi-object} scenarios show a significant drop compared to \emph{single-object} scenarios, it suggests that there is still considerable room for improvement in existing methods for handling the classification of \emph{multi-object} scenarios. \looseness=-1

\begin{wraptable}{r}{0.5\textwidth} 
    \centering
    \vspace{-1em}
    \caption{Performance comparison on the single-object subset.}
    \label{tab:single_object}
    \scalebox{0.9}{ 
        \begin{tabular}{lccc}
        \toprule
            \textbf{Framework} & \texttt{m\_vIoU} & \texttt{tIoU} & \texttt{vIoU@0.5} \\
            \midrule
            \textbf{TubeDETR} & 28.60 & 39.64 & 25.52 \\
            \textbf{TubeDETR-M} & 14.39 & 31.12 & 7.64 \\
            \bottomrule
        \end{tabular}
    }
    \vspace{-2.em}
\end{wraptable}

\paragraph{Comparison within the single-object scenario.}
Section \ref{sec:method} states that TubeDETR-M has additional designs tailored for the new task of DSTVD, but these features make it comparatively less effective than the original TubeDETR in \emph{single-object} scenarios (see Table \ref{tab:single_object}), particularly in \texttt{vIoU} results. The original TubeDETR performs better because it efficiently uses the prior information that one text description corresponds to one object, while TubeDETR-M introduces an additional classification challenge. The difficulty lies in determining whether an object is referenced, making TubeDETR-M harder to optimize. Future work will explore methods to handle multi-object references while maintaining performance in \emph{single-object} scenarios.

\begin{table}[!h]
    \centering
    \vspace{-1.5em}
    \begin{minipage}{0.48\textwidth}
        \centering
        \caption{Performance comparison of description lengths. The variable $l$ denotes the description length, where \emph{short} corresponds to $1 \leq l \leq 5$, \emph{normal} to $6 < l < 10$, and \emph{long} to $l \geq 10$.}
        \vspace{-1.em}
        \scalebox{0.9}{
            \begin{tabular}{lccc}
            \toprule
                \textbf{Length}   & \texttt{m\_vIoU} & \texttt{vIoU@0.5} & \texttt{video-AP@0.25} \\ 
                \midrule
                \emph{short} & 33.64 & 29.61 & 36.29\\
                \emph{normal} & 35.24  & 31.13 & 39.84\\ 
                \emph{long} & 39.53 & 37.26 & 55.02\\ 
                \bottomrule
            \end{tabular}
        }
        \label{tab:description_length}
    \end{minipage} \hfill
    \begin{minipage}{0.48\textwidth}
        \centering
        \caption{Performance comparison of entity counts. The variable $n$ represents the number of entities, where \emph{few} corresponds to  $n = 1$, \emph{moderate} to $2 \leq n \leq 3$, and \emph{many} to $n \geq 4$.}
        \vspace{-1.em}
        \scalebox{0.9}{
            \begin{tabular}{lccc}
            \toprule
                \textbf{Entities}   & \texttt{m\_vIoU} & \texttt{vIoU@0.5} &  \texttt{frame-AP@0.25} \\ 
                \midrule
                \emph{few} & 34.80 & 31.18 & 47.08\\
                \emph{moderate} & 41.31  & 36.77 & 47.41\\ 
                \emph{many} & 19.70 & 22.22 & 21.11\\ 
                \bottomrule
            \end{tabular}
        }
        \label{tab:entity_counts}
    \end{minipage}
    \vspace{-3em}
\end{table}

\paragraph{Comparison across different description lengths.}
Table \ref{tab:description_length} displays the performance of our improved  STCAT-M on the DSTVD task using the DVD-ST dataset. The results indicate that overall, with the increase of description length, our performance in object detection becomes better. The results suggest a positive correlation between the length of the description text and the performance of the STCAT-M. With more information provided in the longer descriptions, tubelet queries, which are generated from the frame embedding and video embedding, can better acquire the position feature and identify the regions in the videos.

\vspace{-1em}
\paragraph{Comparison across different entity counts.}
The results in Table \ref{tab:entity_counts} indicate that STCAT-M performs better on the \emph{moderate} test set compared to \emph{few} and \emph{many}, with \emph{many} being significantly lower than the other two. This suggests that the model's predictive ability is affected when the text descriptions involve either very few or very many entities. In cases of few entities, descriptions may be more ambiguous, while many entities introduce higher semantic complexity, making the model's interpretation more challenging. Additionally, the significant drop in \emph{many} compared to the other two may be attributed to: 1) the relatively limited dataset for this test set, and 2) the need for improvement in the model's text comprehension abilities.



\vspace{-1em}
\section{Conclusion}
\label{sec:conclusion}

In this paper, we introduced the Described Spatial-Temporal Video Detection (DSTVD) benchmark and the DVD-ST dataset, marking a significant advancement in spatial-temporal video understanding. We advance current benchmarks by accommodating a broader range of real-world scenarios, involving more flexible text descriptions and various numbers of referred objects. Moreover, we reformulate the TubeDETR and STCAT models to handle complex, varied text queries and multiple object tracking in video sequences. By enhancing these models and introducing novel elements like tubelet queries and a tubelet-wise matcher, we established a more robust framework for DSTVD. Looking forward, we aim to explore deeper learning architectures and expand our dataset to encompass wider scenarios, driving further innovation in video understanding and its practical applications.

\clearpage  

%
%
\bibliographystyle{splncs04}
\bibliography{main}

\appendix
\newpage
\section{Appendix}
\subsection{Details of the Matching Loss}
In accordance with the definition provided in DETR \cite{detr}, let \( y \) represent the ground truth set of objects, defined as \( y = \{y_{i,j} | i \in [N], j \in [T]\} \), where \( N \) and \( T \) represent the number of objects and time frames, respectively. The prediction set is denoted as \( \hat{y} = \{\hat{y}_{i,j} | i \in [N], j \in [T]\} \), which consists of \( N \times T \) predictions. When \( N \) exceeds the actual number of objects in each video frame, we pad \( y \) to a size of \( N \) with \( \varnothing \) (indicating ``no object'').

The objective is to find a bipartite matching that minimizes the overall cost. This is achieved by finding a permutation \( \sigma \) from the symmetric group \( \mathfrak{S}_N \) that minimizes the following cost function:
\begin{equation}
\hat{\sigma} = \underset{\sigma \in \mathfrak{S}_N}{\arg\min} \sum_{i=1}^{N} \sum_{j=1}^{T} \mathcal{L}_{\text{match}}(y_{i,j}, \hat{y}_{\sigma(i),j})
\end{equation}
where \( \mathcal{L}_{\text{match}}(y_{i,j}, \hat{y}_{\sigma(i),j}) \) represents the matching cost between the ground truth \( y_{i,j} \) and the prediction \( \hat{y}_{\sigma(i),j} \).

The matching cost \( \mathcal{L}_{\text{match}} \) includes both the class prediction and the similarity between predicted and ground truth bounding boxes. For each element \( i \) in the \( j \)-th frame of the ground truth set, we represent it as \( y_{i,j} = (c_{i,j}, b_{i,j}) \), where \( c_{i,j} \) is the target class label, and \( b_{i,j} \in [0,1]^4 \) specifies the bounding box's center coordinates, height, and width, relative to the image size. Correspondingly, for the prediction indexed by \( (\sigma(i), j) \), we define the class probability as \( \hat{p}_{\hat{\sigma}(i),j}(c_{i,j}) \) and the predicted bounding box as \( \hat{b}_{\sigma(i),j} \). The matching cost is defined as:
\begin{equation}
\begin{aligned}
\mathcal{L}_{\text{match}}(y_{i,j}, \hat{y}_{\sigma(i),j}) = -1_{\{c_i \neq \varnothing\}} \log \hat{p}_{\hat{\sigma}(i),j}(c_{i,j}) +  1_{\{c_i \neq \varnothing\}} \mathcal{L}_{\text{box}}(b_{i,j}, \hat{b}_{\sigma(i),j})
\end{aligned}
\end{equation}
This matching strategy is analogous to the heuristic assignment rules in modern object detectors, with the key distinction being the establishment of a one-to-one matching for direct set prediction, avoiding duplicates.

The next step involves calculating the ``Hungarian loss'' for all matched pairs. This loss is a linear combination of a negative log-likelihood for class prediction and a box loss (see DETR \cite{detr} for more details), defined as:
\begin{equation}
\begin{aligned}
\mathcal{L}_{\text {Hungarian }}(y, \hat{y})=\sum_{i=1}^N\sum_{j=1}^{T}\big{[}-\log\hat{p}_{\hat{\sigma}(i),j}\left(c_{i,j}\right)+1_{\left\{c_i \neq \varnothing\right\}}\mathcal{L}_{\text{box}}(b_{i,j},\hat{b}_{\sigma(i),j})\big{]}
\end{aligned}
\end{equation}
where \( \hat{\sigma} \) is the optimal assignment computed in the first step. 

\subsection{Qualitative Results}
\begin{figure*}[!h]
    \includegraphics[width=\textwidth]{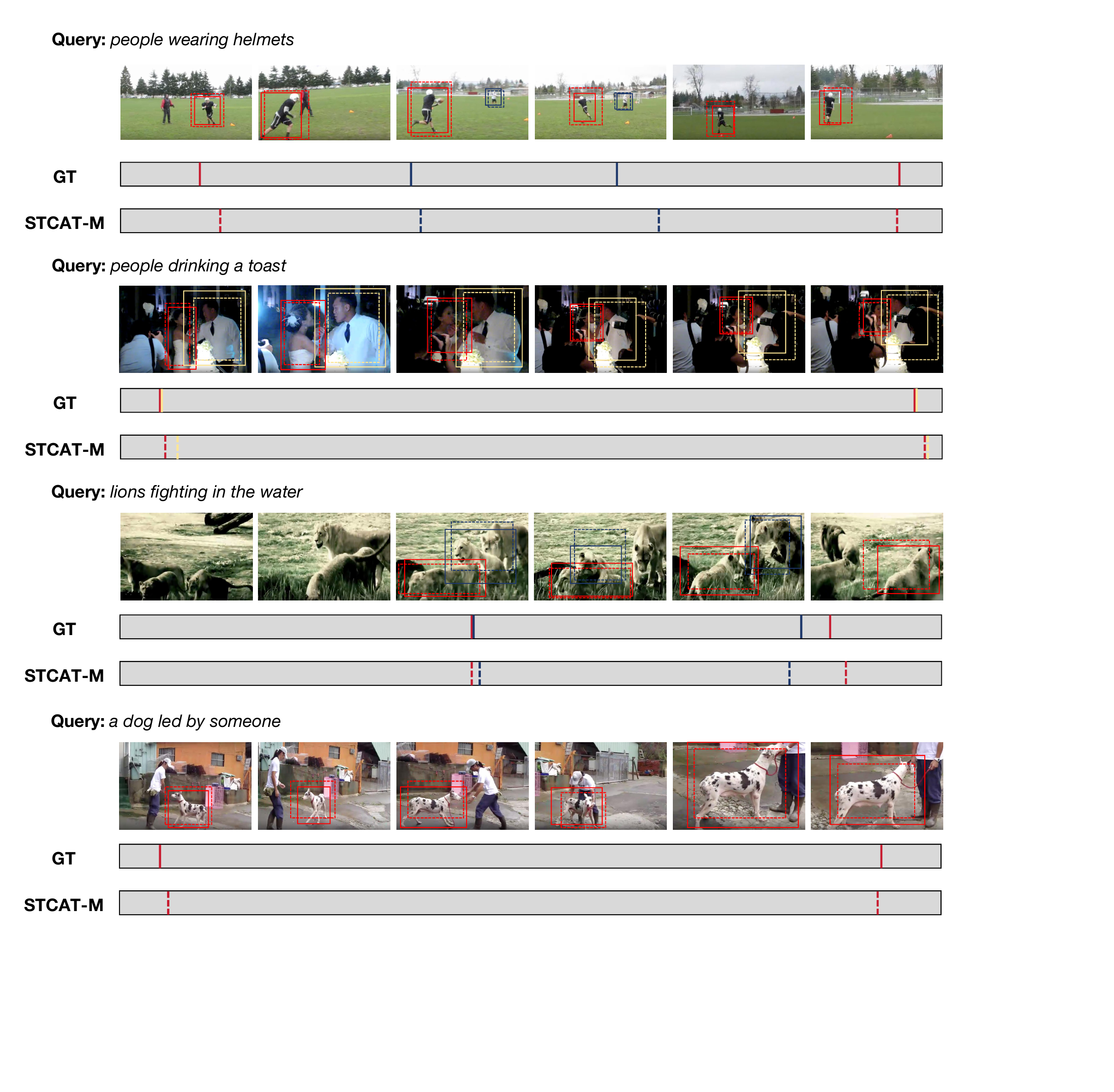}
    \caption{Qualitative examples of spatial-temporal tubelets predicted by STCAT-M, compared with ground truth.}
    \label{fig:qualitative_examples}
\end{figure*}

Figure \ref{fig:qualitative_examples} presents qualitative examples of our predictions on the DVD-ST test set. The comparison with the Ground Truth demonstrates the effectiveness of our methodology.

\end{document}